\documentclass[runningheads]{llncs}

\usepackage{eccv}

\usepackage{eccvabbrv}

\usepackage{graphicx}
\usepackage{booktabs}
\usepackage{amssymb}
\usepackage{amsmath}

\usepackage[accsupp]{axessibility}  

\usepackage{hyperref}

\usepackage{orcidlink}

\begin{document}

\title{BrainRiem: Riemannian Prototype Learning for Source-Free Cross-Site Brain Network Diagnosis}
\titlerunning{BrainRiem: Source-Free Brain Network Diagnosis}

\author{
Kunyu Zhang\inst{1}\textsuperscript{\dag}
\and
Tianxiang Xu\inst{2}\textsuperscript{\dag}
}

\authorrunning{K. Zhang and T. Xu}

\institute{
Zhengzhou University, Zhengzhou 450000, China\\
\email{kunyu.zky@gmail.com}
\and
Peking University, Beijing 102600, China
}

\maketitle
\begingroup
\renewcommand{\thefootnote}{\dag}
\footnotetext{The authors contributed equally to this research.}
\endgroup

\begin{abstract}
Multi-site functional MRI (fMRI) studies are essential for robust neuropsychiatric diagnosis yet suffer severe domain shifts from scanner heterogeneity, demographics, and site-specific acquisition protocols. Traditional domain adaptation requires concurrent source and target data access, violating clinical privacy regulations. Moreover, functional connectivity matrices lie on the Symmetric Positive Definite (SPD) manifold, where Euclidean operations cause geometric distortions corrupting diagnostic patterns. We propose BrainRiem, a source-free domain adaptation framework learning compact Riemannian brain prototypes via manifold-aware bi-level optimization. It employs the Log-Euclidean Metric to ensure prototypes remain valid SPD matrices, while Dirichlet Energy spectral calibration aligns their frequency characteristics with real brain networks. Only anonymized prototypes are transmitted to target sites, serving as stable anchors for training local models without source data access and reducing leakage under the evaluated attacks. Comprehensive experiments on ABIDE and REST-meta-MDD show BrainRiem consistently outperforms state-of-the-art source-free, traditional, and graph domain adaptation methods across diverse scanners and demographics. Notably, learned prototypes exhibit biologically interpretable connectivity patterns aligning with established neuroscience findings, validating the necessity of Riemannian geometry for brain network analysis.

\end{abstract}

\section{Introduction}
\begin{figure*}[!t]
    \centering
    \includegraphics[width=0.95\textwidth]{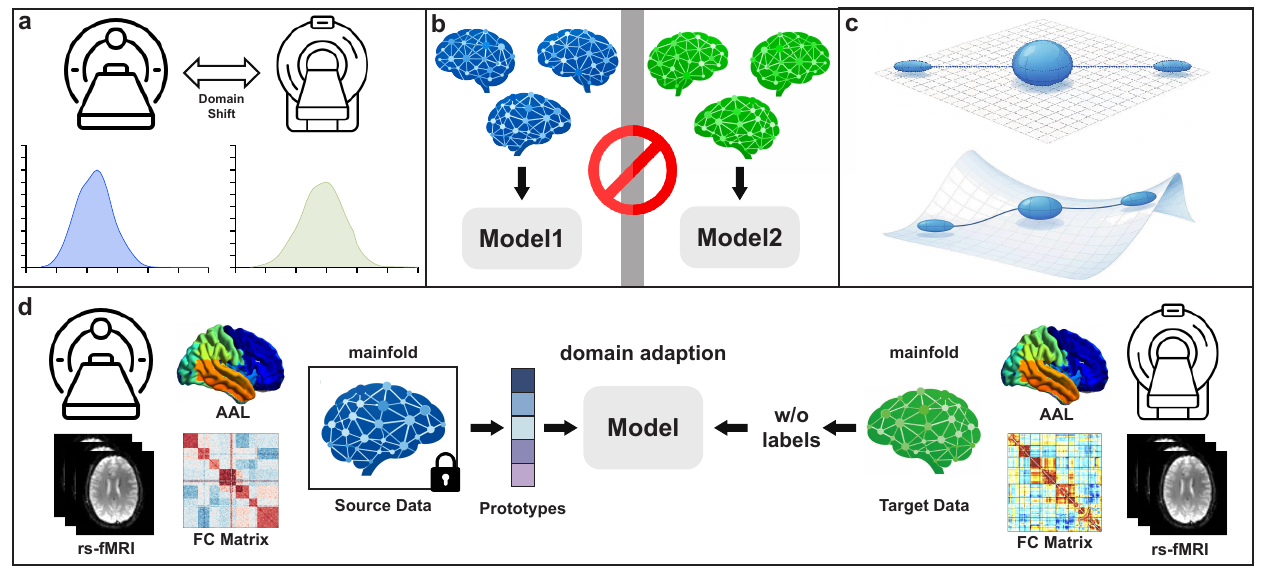}
    \caption{Motivation for BrainRiem. \textbf{(a)} Domain shift from scanner heterogeneity degrades model performance. \textbf{(b)} Data sharing constraints in multi-site collaborations. \textbf{(c)} Euclidean operations distort SPD manifold geometry; Riemannian operations preserve structure. \textbf{(d)} BrainRiem learns manifold-aware prototypes for source-free adaptation.}
    \label{fig:motivation}
\end{figure*}

Autism Spectrum Disorder (ASD) and Major Depressive Disorder (MDD) are pervasive psychiatric conditions imposing substantial socioeconomic burdens. While early diagnosis is essential for intervention, traditional clinical evaluations rely heavily on subjective behavioral observations. Recently, deep learning on resting-state functional magnetic resonance imaging (rs-fMRI) has emerged as a promising avenue for objective diagnosis~\cite{brainmass,zhang2025mvho}. These methods utilize functional connectivity (FC) patterns to identify neurobiological signatures of psychiatric disorders. Furthermore, large-scale multi-site datasets, such as ABIDE~\cite{abide} for ASD and REST-meta-MDD~\cite{restmeatmdd} for MDD, facilitate developing robust and generalizable models.

Despite these advancements, deploying deep learning models across multiple sites poses three fundamental challenges (Figure~\ref{fig:motivation}a--c). First, models trained on multi-site data often exhibit severe degradation on unseen sites. Scanner heterogeneity, including variations in manufacturers, magnetic field strengths, and acquisition protocols, introduces systematic biases confounding disease signals. This domain shift necessitates domain adaptation for effective knowledge transfer~\cite{wang2026brain}. Second, conventional methods require concurrent source and target data access. However, sharing raw neuroimaging data is often infeasible due to regulations (e.g., GDPR, HIPAA) and logistical barriers. This motivates Source-Free Domain Adaptation (SFDA) to transfer pre-trained models without source data, providing basic data protection~\cite{wang2026usbd}. Third, existing methods predominantly operate in Euclidean space, neglecting the intrinsic geometry of functional connectivity (FC) matrices. As Symmetric Positive Definite (SPD) matrices, they naturally reside on a Riemannian manifold~\cite{wang2026riemannian}. Standard Euclidean operations violate positive definiteness and induce the ``swelling effect,'' inflating determinants and yielding biologically implausible representations. This geometric mismatch underscores the necessity for manifold-aware methods preserving the SPD structure.

Although existing approaches address these challenges in isolation, they fail to provide a unified solution. Traditional domain adaptation~\cite{ganin2016domain, sun2016deep, tzeng2017adversarial,dann} aligns feature distributions via adversarial training or moment matching but necessitates concurrent source and target data access. Source-free methods~\cite{liang2020we, rahman20213c,wang2026usbd,ding2022source} circumvent privacy constraints by transferring pre-trained models yet predominantly operate in Euclidean space, causing geometric distortions for Symmetric Positive Definite (SPD) data. Recent graph OOD/domain generalization studies further highlight the importance of structural and geometric robustness~\cite{wang2026riemannian,liu2026information,gifford20267t}. Conversely, manifold-aware approaches explicitly account for the Riemannian structure of functional connectivity matrices but generally require source data access. To our knowledge, prior source-free cross-site fMRI methods have not jointly addressed transferable SPD-valid prototypes, cross-site generalization, and empirical leakage under explicit attacks.

To address these limitations, we propose BrainRiem, a manifold-aware Source-Free Domain Adaptation (SFDA) framework utilizing Learnable Brain Prototypes (Figure~\ref{fig:motivation}d). Our premise is that compact, geometry-preserving prototypes serve as effective knowledge carriers guiding target adaptation without raw source data. Specifically, we integrate Riemannian constraints via the Log-Euclidean Metric (LEM)~\cite{arsigny2007geometric} into a bi-level optimization scheme, ensuring prototypes adhere to the Symmetric Positive Definite (SPD) manifold geometry. Furthermore, to mitigate overfitting scanner artifacts, we introduce a Dirichlet energy-based spectral calibration, constraining prototypes to preserve intrinsic source brain network frequency characteristics. By transferring compact prototypes alongside the pre-trained model instead of raw data, our framework reduces privacy risks from statistical alignment while enabling effective cross-site knowledge transfer.

We validate BrainRiem on ABIDE and REST-meta-MDD through leave-one-site-out experiments. Results demonstrate superior accuracy compared to both Euclidean source-free methods and manifold-aware methods requiring source access. Our contributions are as follows:
\begin{itemize}
    \item We introduce a manifold-aware source-free domain adaptation framework for brain networks, combining geometric structure preservation with reduced data sharing requirements through learnable brain prototypes.
    \item We identify and formalize the \textit{geometric drift} problem in source-free adaptation: models trained in Euclidean space suffer manifold distortion when transferred without source anchors. Our bi-level optimization framework embeds Riemannian constraints to prevent this drift, with Dirichlet Energy providing spectral calibration for scanner-induced variations.
    \item We provide extensive validation on two large-scale multi-site benchmarks, demonstrating that manifold-aware prototype learning significantly outperforms Euclidean alternatives, establishing a new paradigm for practical multi-site collaborative neuroimaging.
\end{itemize}

\section{Related Work}

\subsection{Graph-Based fMRI Analysis}
Graph-based modeling is widely used in fMRI analysis, where ROIs are represented as nodes and functional connectivity is represented as weighted edges. GNN models, including GCN~\cite{gcn}, GraphSAGE~\cite{graphsage}, BrainNetCNN~\cite{kawahara2017brainnetcnn}, and BrainGNN~\cite{li2021braingnn}, have demonstrated strong performance in brain disorder diagnosis by modeling inter-regional dependencies~\cite{parisot2017spectral,tang2024contrastive,zhang2026modeling}. Recent studies further explored large-scale self-supervision~\cite{brainmass}, cross-modal mutual learning~\cite{10182318}, and topology regularization~\cite{yang2024topology}.

In parallel, Riemannian methods for SPD representations have been introduced to preserve the geometry of covariance and connectivity descriptors. The Log-Euclidean framework~\cite{arsigny2007geometric}, SPDNet~\cite{huang2017spdnet}, and covariance-based methods~\cite{acharya2018covariance} reduce the distortion introduced by naive Euclidean operations~\cite{brooks2019riemannian,ju2023graph}. However, most of these studies focus on in-domain learning, and limited attention has been given to source-free cross-site transfer.

\subsection{Source-Free Domain Adaptation on Graphs}
Cross-site fMRI diagnosis is strongly affected by domain shifts induced by scanner variation and demographic heterogeneity. Graph domain adaptation methods, such as UDA-GCN~\cite{wu2020unsupervised}, align source and target representations, but they usually require joint access to both domains during training~\cite{liu2023structural,you2023graph,wang2026nested}. Such a requirement is often impractical in medical collaboration settings with strict privacy constraints.

Source-free domain adaptation removes source-data sharing and performs adaptation with a pre-trained source model and unlabeled target data~\cite{kundu2021generalize,dong2021confident}. Representative SFDA methods, including SHOT~\cite{liang2020we}, NRC~\cite{yang2021exploiting}, and 3C-GAN~\cite{rahman20213c}, are mainly developed in Euclidean feature spaces. Recent graph studies have explored universal structural distillation, Riemannian flow matching, and information-bottleneck structural learning~\cite{wang2026usbd,wang2026riemannian,liu2026information,wang2026sgac}. For brain connectivity data with SPD constraints, this design can weaken geometric fidelity and limit adaptation robustness~\cite{jiang2022motor}.
Broader neuroimaging work has also begun to examine out-of-distribution modeling from dedicated fMRI datasets~\cite{gifford20267t}, further motivating robust cross-site adaptation.

\subsection{Prototype Learning}
Prototype learning summarizes source-domain knowledge into compact class anchors. Prototypical Networks~\cite{snell2017prototypical} established this paradigm, and subsequent transfer methods adopted prototypes as stable anchors for alignment under distribution shift~\cite{sung2018learning,liu2020prototype,li2021adaptive,zhang2021rethinking}. Recent brain-network studies also combined prototypes with interpretable reasoning mechanisms~\cite{zhang2026pime}.

In privacy-sensitive neuroimaging collaboration, prototypes are practical because compact summaries can replace raw data exchange. However, most prototype construction strategies rely on Euclidean averaging or clustering, which can violate SPD geometry. This limitation motivates manifold-aware prototype learning with Riemannian-consistent operations.

\section{Methodology}
\begin{figure*}[!t]
    \centering
    \includegraphics[width=0.9\textwidth]{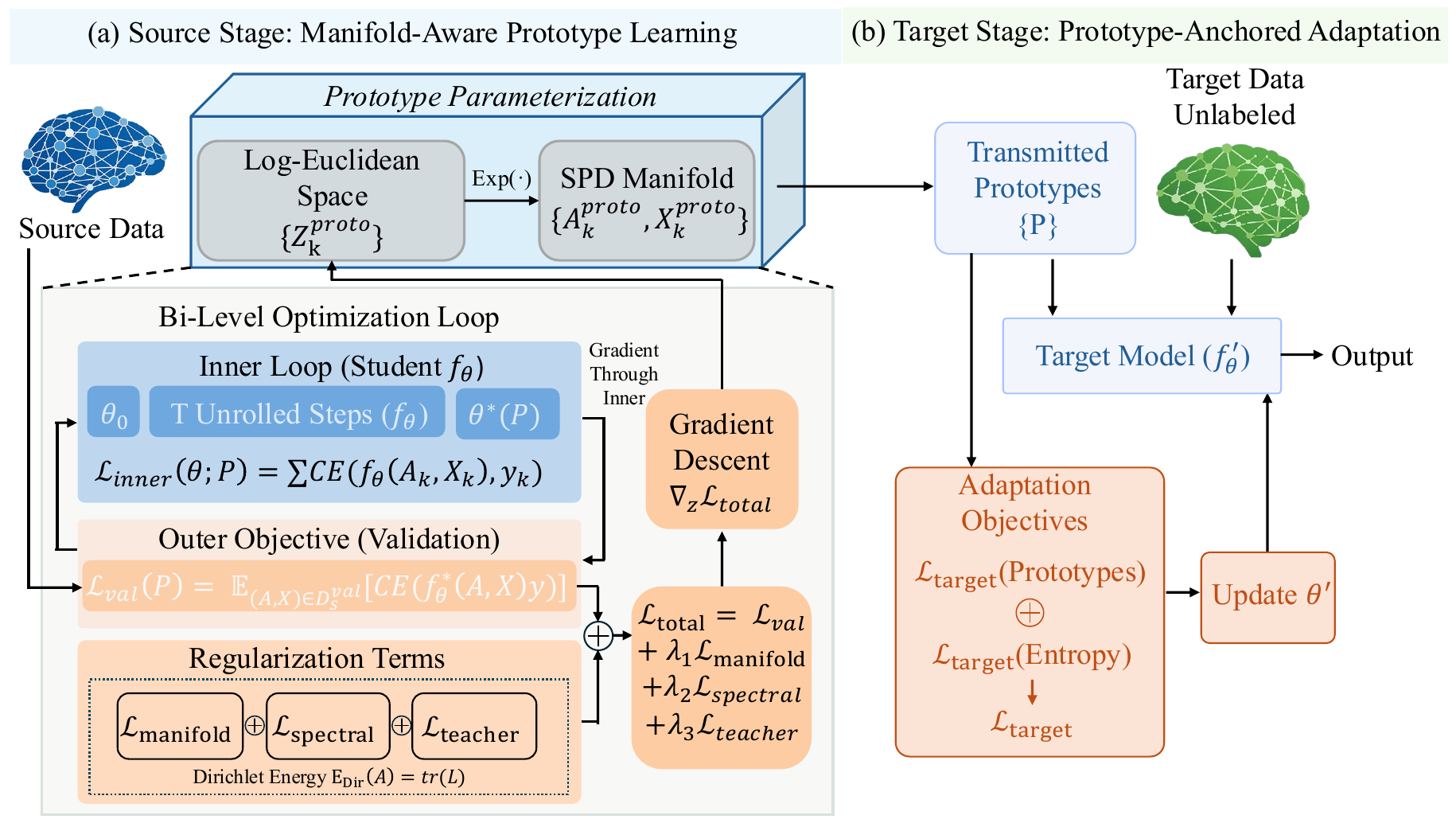}
    \caption{Overview of the BrainRiem framework. The method consists of two privacy-compliant stages: (a) Source-site prototype learning via manifold-aware bi-level optimization; (b) Target-site adaptation using transmitted prototypes as stable anchors.}
    \label{fig:framework}
\end{figure*}

The framework in Figure~\ref{fig:framework} has two stages. In source-stage learning, prototypes are optimized on the SPD manifold via bi-level optimization with geometric and diagnostic regularization. In target-stage adaptation, only $\mathcal{P}$ is shared and used as supervised anchors together with entropy minimization on unlabeled target data.

\subsection{Problem Formulation and Preliminaries}

Let $\mathcal{S}_{++}^N = \{\mathbf{A} \in \mathbb{R}^{N \times N}: \mathbf{A}=\mathbf{A}^\top, \mathbf{A}\succ 0\}$ denote the SPD manifold. Each subject is represented as $\mathcal{G}=(\mathbf{A},\mathbf{X})$, where $\mathbf{A}\in\mathcal{S}_{++}^N$ is the functional connectivity matrix and $\mathbf{X}=\mathbf{I}_N$ encodes ROI identities. In source-free adaptation, labeled source data $\mathcal{D}_s$ are used only at source stage, and only unlabeled target data $\mathcal{D}_t$ are available during adaptation.

We learn a compact prototype set
\begin{equation}
\mathcal{P}=\{(\mathbf{A}_k^{proto},\mathbf{X}_k^{proto},y_k)\}_{k=1}^{K}, \quad K \ll |\mathcal{D}_s|,
\end{equation}
which is transmitted instead of raw source samples.

\subsection{Manifold-Aware Prototype Parameterization}

We parameterize prototypes in Log-Euclidean tangent space and map them back to the SPD manifold:
\begin{equation}
\mathbf{A}_k^{proto}=\mathrm{Exp}(\mathbf{Z}_k^{proto}), \quad \mathbf{Z}_k^{proto}\in\mathbb{R}^{N\times N}_{sym}.
\end{equation}
\noindent Detailed definitions of Log/Exp mappings and the Log-Euclidean distance are provided in Supplementary Material A.1.
\textbf{Design choice.} Prototype optimization is performed on $\{\mathbf{Z}_k^{proto}\}$ to keep Euclidean optimization simple while enforcing valid SPD prototypes through $\mathrm{Exp}(\cdot)$.

\textbf{Optimization role.} This parameterization enforces manifold validity of prototypes during all optimization steps.

\subsection{Bi-Level Optimization for Knowledge Transfer}

\textbf{Assumption.} Prototype quality is measured by the performance of a student model trained only on prototypes and evaluated on held-out source validation data.

The inner problem trains the student model $f_\theta$ on prototypes:
\begin{equation}
\mathcal{L}_{inner}(\theta;\mathcal{P})=\sum_{(\mathbf{A}_k^{proto},\mathbf{X}_k^{proto},y_k)\in\mathcal{P}}\ell_{CE}(f_\theta(\mathbf{A}_k^{proto},\mathbf{X}_k^{proto}),y_k),
\end{equation}
\begin{equation}
\theta^*(\mathcal{P})=\arg\min_{\theta}\mathcal{L}_{inner}(\theta;\mathcal{P}).
\end{equation}

The outer objective evaluates $f_{\theta^*(\mathcal{P})}$ on source validation data:
\begin{equation}
\mathcal{L}_{val}(\mathcal{P})=\frac{1}{|\mathcal{D}_s^{val}|}\sum_{(\mathbf{A},\mathbf{X},y)\in\mathcal{D}_s^{val}}\ell_{CE}(f_{\theta^*(\mathcal{P})}(\mathbf{A},\mathbf{X}),y).
\end{equation}

We use a regularized outer objective:
\begin{equation}
\mathcal{L}_{total}=\mathcal{L}_{val}+\lambda_1\mathcal{L}_{manifold}+\lambda_2\mathcal{L}_{spectral}+\lambda_3\mathcal{L}_{teacher},
\min_{\{\mathbf{Z}_k^{proto}\}}\mathcal{L}_{total}.
\end{equation}
\textbf{Optimization role.} $\mathcal{L}_{val}$ optimizes transferability, while regularization terms control geometry, spectral statistics, and diagnostic consistency.

\textbf{First-Order Approximation.}
Gradients through the inner optimization are computed with a first-order approximation using $T$ unrolled inner steps~\cite{finn2017model,nichol2018first}.
Detailed gradient derivations are provided in Supplementary Material A.2.
In each source-stage iteration, we first run $T$ inner updates for $f_\theta$ on the current prototypes, then update prototype parameters $\{\mathbf{Z}_k^{proto}\}$ by descending $\nabla_{\mathbf{Z}}\mathcal{L}_{total}$, while keeping source data fixed.

\subsection{Spectral Calibration via Dirichlet Energy}

Define graph Laplacian $\mathbf{L}=\mathbf{D}-\mathbf{A}$. Dirichlet Energy is
\begin{equation}
E_{Dir}(\mathbf{A},\mathbf{X})=\mathrm{tr}(\mathbf{X}^\top\mathbf{L}\mathbf{X}).
\end{equation}
In our setting, $\mathbf{X}=\mathbf{I}_N$, so $E_{Dir}(\mathbf{A})=\mathrm{tr}(\mathbf{L})$.

We regularize source-prototype spectral statistics by
\begin{equation}
\mathcal{L}_{spectral}=\sum_{k=1}^{K}\left\|E_{Dir}(\mathbf{A}_k^{proto},\mathbf{I}_N)-\mathbb{E}_{\mathbf{A}^s\in\mathcal{D}_s}[E_{Dir}(\mathbf{A}^s,\mathbf{I}_N)]\right\|^2.
\label{eq:spectral}
\end{equation}
Additional mathematical details of this spectral term are provided in Supplementary Material C.3.
\textbf{Optimization role.} This term aligns prototype-level spectral characteristics with source-domain statistics.

\subsection{Diagnostic Consistency and Geometry Preservation}

Teacher-guided supervision is
\begin{equation}
\mathcal{L}_{teacher}=\sum_{k=1}^{K}\ell_{CE}(y_k,p_t(\mathbf{A}_k^{proto},\mathbf{X}_k^{proto})).
\end{equation}
\textbf{Optimization role.} This term preserves class-discriminative semantics in prototypes.

Manifold consistency is
\begin{equation}
\mathcal{L}_{manifold}=\sum_{k=1}^{K}\sum_{i\in\mathcal{N}_k}\left\|d_{LE}(\mathbf{A}_k^{proto},\mathbf{A}_i^s)-d_0(k,i)\right\|^2.
\end{equation}
\noindent Here, $\mathcal{N}_k$ denotes the $k$-th prototype neighborhood built by $m$-nearest source samples under $d_{LE}$, and $d_0(k,i)$ is the corresponding initial Log-Euclidean distance before prototype refinement.
\textbf{Optimization role.} This term preserves local geodesic structure around prototypes.

\subsection{Deployment Protocol}

\textbf{Source Stage:} Optimize $\mathcal{P}$ from $\mathcal{D}_s$ by minimizing $\mathcal{L}_{total}$.
\textbf{Transmission:} Share only $\mathcal{P}$.
\textbf{Target Stage:} Adapt a local model $f_{\theta'}$ using prototype supervision and target entropy minimization:
\begin{equation}
\begin{aligned}
\mathcal{L}_{target}
&=\sum_{(\mathbf{A}_k^{proto},\mathbf{X}_k^{proto},y_k)\in\mathcal{P}}
\ell_{CE}\big(f_{\theta'}(\mathbf{A}_k^{proto},\mathbf{X}_k^{proto}),y_k\big) \\
&\quad - \lambda_4\mathbb{E}_{(\mathbf{A}^t,\mathbf{X}^t)\in\mathcal{D}_t}
\sum_c p_{\theta'}(c|\mathbf{A}^t,\mathbf{X}^t)\log p_{\theta'}(c|\mathbf{A}^t,\mathbf{X}^t).
\end{aligned}
\end{equation}
During target-stage adaptation, prototypes and the teacher model are fixed, and only $\theta'$ is updated.

\section{Experimental Setup and Results}
\begin{figure*}[!t]
    \centering
    \includegraphics[width=\textwidth]{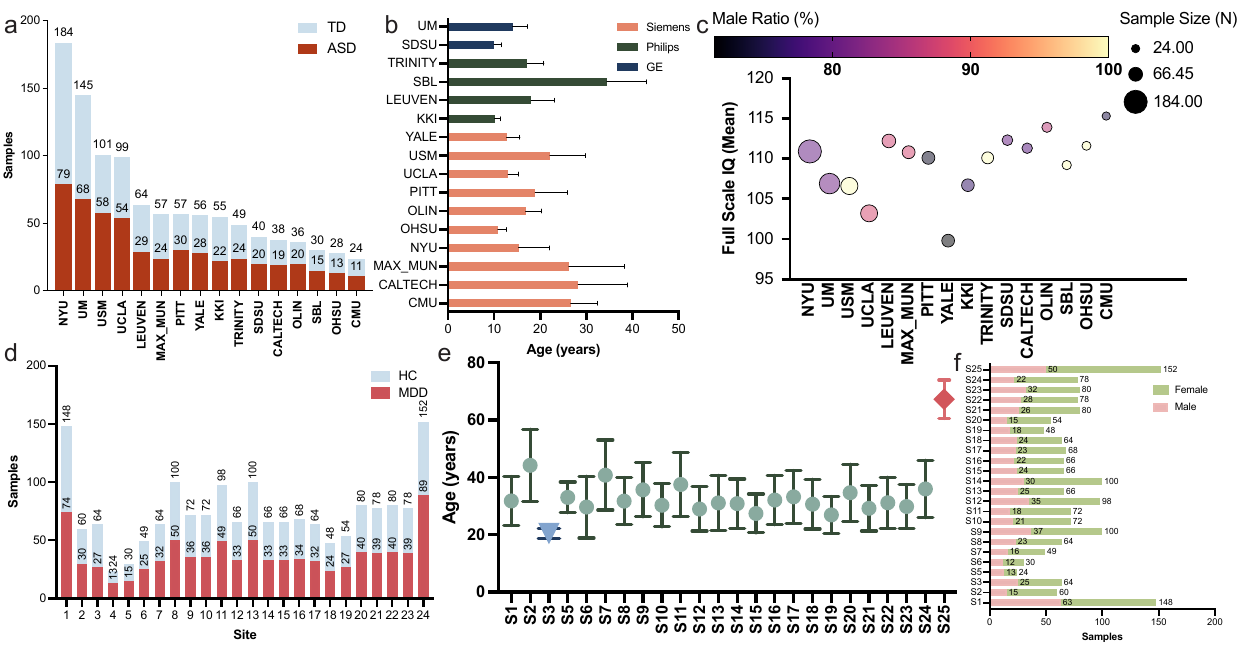}
    \caption{Statistical heterogeneity across multi-site datasets. ABIDE \textbf{(a-c)}: (a) Sample size imbalance across sites. (b) Age distribution by scanner vendor (Siemens, Philips, GE). (c) Demographic diversity (IQ, gender ratio). REST-meta-MDD \textbf{(d-f)}: (d) Class balance variations. (e) Extreme age shifts (S3: $\sim$20yrs vs. S25: $\sim$67yrs). (f) Gender distribution biases.}
    \label{fig:dataset}
\end{figure*}
\subsection{Experimental Settings}
\textbf{Datasets and Preprocessing.} We evaluated our framework on two large-scale multi-site fMRI datasets: the Autism Brain Imaging Data Exchange (ABIDE) and the REST-meta-MDD consortium dataset. As illustrated in Figure~\ref{fig:dataset}, both datasets exhibit significant inter-site heterogeneity in sample size, age, gender, and scanner configurations, confirming that source-free domain adaptation is strictly necessary. During quality control, Yale in ABIDE and S20 in REST-meta-MDD were excluded before model training and evaluation because their computed FC-matrix files were incomplete due to a storage issue. All reported results therefore use the remaining 15 and 23 sites, respectively. Standard fMRI preprocessing was applied to construct $116 \times 116$ functional connectivity matrices (see Supplementary Material B.1 for complete protocols and statistical characterization). 

\textbf{Adaptation Scenarios.} We designed two cross-site adaptation scenarios. In Single-Source Adaptation, we select one large site as the labeled source and adapt to each remaining site individually, testing the method's ability to transfer knowledge from a single distribution to diverse targets. In Leave-One-Site-Out (LOSO) Adaptation, we hold out one site as the unlabeled target and combine all other sites as the labeled source, mimicking a federated-like setting.

\textbf{Protocol and Data Partitioning.}We adopt a strictly Inductive Source-Free Domain Adaptation setting with subject-wise partitioning into disjoint Adaptation and Test sets. For LOSO, the source model is trained using all sites except target $S_t$, with site-balanced sampling. For Single-Source ($S_i \rightarrow S_j$), the source model uses 5-fold cross-validation. All hyperparameters were selected once using source-only validation on the training source domain(s), then fixed for all target sites and all adaptation tasks. No target-domain labels or target-specific model selection were used.

\begin{table*}[!t]
\centering
\caption{Single-source adaptation on ABIDE. Each column: one source to all targets. Bold: best.}
\label{tab:abide_single_source}
\resizebox{\textwidth}{!}{%
\begin{tabular}{l|cccc|cccc|cccc|cccc|cccc}
\toprule
& \multicolumn{4}{c|}{\textbf{NYU$\rightarrow$}} & \multicolumn{4}{c|}{\textbf{LEUVEN$\rightarrow$}} & \multicolumn{4}{c|}{\textbf{UCLA$\rightarrow$}} & \multicolumn{4}{c|}{\textbf{UM$\rightarrow$}} & \multicolumn{4}{c}{\textbf{USM$\rightarrow$}} \\
\textbf{Method} & LEU & UCLA & UM & USM & NYU & UCLA & UM & USM & NYU & LEU & UM & USM & NYU & LEU & UCLA & USM & NYU & LEU & UCLA & UM \\
\midrule
SHOT & 62.2 & 63.3 & 58.9 & 62.2 & 60.3 & 62.9 & 59.5 & 62.1 & 59.5 & 62.8 & 59.3 & 62.2 & 59.8 & 62.3 & 63.2 & 61.1 & 59.7 & 62.8 & 62.7 & 58.9 \\
NRC & 63.6 & 64.2 & 60.5 & 63.1 & 61.3 & 64.0 & 59.9 & 64.6 & 60.9 & 65.1 & 60.5 & 63.4 & 61.0 & 64.3 & 64.7 & 63.3 & 61.0 & 64.6 & 64.9 & 60.7 \\
3C-GAN & 61.0 & 61.3 & 58.1 & 59.8 & 58.9 & 60.1 & 58.1 & 59.7 & 58.1 & 60.7 & 58.1 & 60.7 & 59.1 & 61.5 & 60.5 & 60.2 & 58.9 & 59.7 & 60.4 & 58.4 \\
G-SFDA & 62.6 & 62.1 & 58.7 & 62.3 & 59.7 & 63.3 & 59.3 & 62.5 & 59.7 & 62.2 & 59.7 & 62.8 & 59.3 & 62.6 & 63.1 & 62.1 & 59.5 & 63.2 & 62.2 & 59.1 \\
\midrule
DANN & 60.1 & 61.7 & 58.7 & 60.7 & 58.6 & 61.6 & 58.0 & 59.6 & 58.6 & 61.2 & 58.1 & 60.7 & 58.1 & 60.4 & 60.0 & 60.3 & 59.0 & 61.3 & 61.3 & 58.2 \\
CORAL & 56.9 & 58.2 & 57.2 & 57.5 & 58.0 & 58.3 & 56.4 & 57.0 & 56.4 & 57.7 & 56.6 & 59.1 & 57.5 & 58.5 & 59.1 & 56.9 & 57.5 & 57.0 & 58.3 & 57.5 \\
\midrule
SPDNet & 50.4 & 46.8 & 51.4 & 47.5 & 50.1 & 49.8 & 52.5 & 47.6 & 50.4 & 47.7 & 50.9 & 48.8 & 52.0 & 48.5 & 47.9 & 51.1 & 51.6 & 50.6 & 49.8 & 52.6 \\
SPD-GNN & 49.1 & 47.7 & 50.9 & 50.7 & 51.8 & 48.7 & 52.4 & 47.7 & 51.7 & 48.6 & 50.0 & 48.1 & 51.0 & 48.9 & 46.6 & 49.9 & 49.7 & 47.3 & 49.9 & 51.9 \\
\midrule
UDA-GCN & 63.8 & 64.7 & 60.0 & 63.9 & 61.7 & 64.5 & 60.6 & 63.5 & 61.0 & 64.2 & 60.3 & 63.4 & 60.9 & 64.3 & 64.1 & 63.7 & 61.2 & 64.2 & 64.3 & 60.5 \\
StruRW & 64.5 & 64.2 & 60.4 & 63.3 & 61.5 & 65.0 & 60.4 & 64.1 & 60.6 & 65.0 & 60.9 & 64.1 & 61.1 & 64.8 & 64.0 & 63.2 & 61.2 & 64.0 & 64.4 & 60.5 \\
\midrule
BrainGNN & 55.7 & 55.0 & 55.8 & 54.2 & 53.9 & 53.9 & 54.9 & 53.6 & 55.2 & 54.6 & 54.3 & 56.1 & 55.0 & 53.4 & 54.5 & 55.9 & 54.5 & 55.7 & 55.5 & 54.3 \\
A-GCL & 58.1 & 59.2 & 56.2 & 57.8 & 57.0 & 58.9 & 57.4 & 58.0 & 56.3 & 58.2 & 56.5 & 58.7 & 57.7 & 58.8 & 57.7 & 58.9 & 56.6 & 59.1 & 59.0 & 56.2 \\
\midrule
\textbf{Ours} & \textbf{70.0} & \textbf{71.1} & \textbf{63.7} & \textbf{69.4} & \textbf{66.0} & \textbf{70.8} & \textbf{63.8} & \textbf{70.1} & \textbf{64.6} & \textbf{70.8} & \textbf{64.5} & \textbf{69.8} & \textbf{64.8} & \textbf{71.1} & \textbf{71.0} & \textbf{69.0} & \textbf{65.2} & \textbf{70.6} & \textbf{70.7} & \textbf{64.2} \\
\bottomrule
\end{tabular}%
}
\end{table*}

\begin{table*}[!t]
\centering
\caption{Single-source adaptation on REST-meta-MDD. Each column: one source to all targets. Bold: best.}
\label{tab:mdd_single_source}
\resizebox{\textwidth}{!}{%
\begin{tabular}{l|cccc|cccc|cccc|cccc|cccc}
\toprule
& \multicolumn{4}{c|}{\textbf{S1$\rightarrow$}} & \multicolumn{4}{c|}{\textbf{S9$\rightarrow$}} & \multicolumn{4}{c|}{\textbf{S12$\rightarrow$}} & \multicolumn{4}{c|}{\textbf{S14$\rightarrow$}} & \multicolumn{4}{c}{\textbf{S25$\rightarrow$}} \\
\textbf{Method} & S9 & S12 & S14 & S25 & S1 & S12 & S14 & S25 & S1 & S9 & S14 & S25 & S1 & S9 & S12 & S25 & S1 & S9 & S12 & S14 \\
\midrule
SHOT & 59.9 & 57.7 & 60.7 & 58.3 & 58.5 & 57.8 & 62.0 & 60.0 & 58.9 & 59.9 & 62.4 & 58.8 & 59.1 & 60.9 & 57.7 & 59.3 & 58.4 & 59.9 & 56.6 & 60.2 \\
NRC & 60.5 & 58.2 & 62.1 & 59.8 & 59.4 & 58.4 & 63.0 & 60.8 & 59.4 & 61.2 & 63.0 & 60.1 & 60.2 & 62.5 & 58.0 & 60.5 & 59.1 & 60.7 & 57.7 & 62.5 \\
3C-GAN & 58.7 & 56.9 & 59.3 & 57.5 & 57.3 & 57.1 & 59.6 & 58.5 & 57.6 & 59.0 & 60.4 & 57.9 & 57.7 & 59.3 & 56.6 & 57.8 & 57.4 & 58.5 & 56.6 & 59.2 \\
G-SFDA & 59.3 & 57.5 & 60.3 & 58.3 & 58.4 & 57.7 & 61.8 & 59.7 & 58.2 & 59.6 & 61.1 & 59.3 & 59.0 & 60.9 & 57.5 & 59.2 & 58.2 & 59.8 & 57.1 & 61.2 \\
\midrule
DANN & 58.1 & 56.9 & 59.5 & 58.0 & 57.7 & 57.4 & 60.0 & 58.7 & 57.4 & 58.5 & 60.9 & 58.0 & 58.3 & 59.3 & 56.6 & 58.3 & 57.6 & 58.2 & 56.5 & 58.5 \\
CORAL & 56.4 & 55.8 & 56.4 & 56.5 & 56.3 & 56.0 & 58.2 & 57.5 & 56.0 & 56.4 & 58.0 & 56.9 & 57.0 & 57.7 & 56.0 & 56.1 & 56.4 & 56.2 & 55.7 & 57.3 \\
\midrule
SPDNet & 50.8 & 53.6 & 49.2 & 52.8 & 52.8 & 52.9 & 50.0 & 51.2 & 52.1 & 50.4 & 50.4 & 51.0 & 51.5 & 50.3 & 53.5 & 50.9 & 51.1 & 51.6 & 53.1 & 51.0 \\
SPD-GNN & 51.1 & 53.3 & 49.3 & 51.1 & 51.9 & 52.7 & 49.0 & 50.1 & 52.0 & 51.1 & 48.4 & 51.3 & 50.4 & 49.6 & 53.2 & 50.4 & 51.2 & 52.3 & 52.9 & 49.4 \\
\midrule
UDA-GCN & 60.9 & 58.1 & 61.6 & 59.4 & 59.2 & 58.5 & 62.6 & 60.8 & 59.6 & 61.6 & 64.0 & 60.2 & 59.9 & 62.0 & 58.2 & 60.7 & 59.4 & 61.2 & 57.3 & 62.4 \\
StruRW & 60.7 & 58.0 & 61.7 & 59.8 & 59.3 & 58.5 & 62.9 & 60.5 & 59.6 & 61.7 & 63.7 & 60.5 & 60.0 & 62.0 & 57.9 & 60.8 & 59.5 & 60.8 & 57.7 & 62.4 \\
\midrule
BrainGNN & 55.3 & 54.7 & 54.0 & 55.1 & 55.6 & 54.5 & 55.7 & 55.7 & 55.0 & 55.8 & 56.3 & 55.9 & 55.3 & 54.7 & 54.9 & 55.7 & 54.7 & 55.5 & 54.9 & 54.7 \\
A-GCL & 56.3 & 55.9 & 57.1 & 56.9 & 56.2 & 56.0 & 57.1 & 56.2 & 56.8 & 56.9 & 57.6 & 57.1 & 56.4 & 56.7 & 56.3 & 56.4 & 55.9 & 57.2 & 55.6 & 57.5 \\
\midrule
\textbf{Ours} & \textbf{64.5} & \textbf{60.1} & \textbf{66.2} & \textbf{62.5} & \textbf{62.1} & \textbf{60.8} & \textbf{68.2} & \textbf{64.5} & \textbf{62.4} & \textbf{65.5} & \textbf{69.1} & \textbf{63.8} & \textbf{63.5} & \textbf{67.4} & \textbf{60.1} & \textbf{64.2} & \textbf{62.2} & \textbf{65.1} & \textbf{59.5} & \textbf{66.8} \\
\bottomrule
\end{tabular}%
}
\end{table*}

\begin{table*}[!t]
\centering
\caption{LOSO adaptation results (Accuracy \%). Representative sites shown (LEU=LEUVEN). Avg is computed over \textbf{all} sites (15 for ABIDE, 23 for MDD). See Appendix B.5 for complete results.}
\label{tab:loso}
\resizebox{0.8\textwidth}{!}{%
\begin{tabular}{l|cccccc|cccccc}
\toprule
& \multicolumn{6}{c|}{\textbf{ABIDE}} & \multicolumn{6}{c}{\textbf{REST-meta-MDD}} \\
\textbf{Method} & NYU & LEU & UCLA & UM & USM & Avg & S1 & S9 & S12 & S14 & S25 & Avg \\
\midrule
SHOT & 65.5 & 68.9 & 70.3 & 63.1 & 69.9 & 68.5 & 61.1 & 64.9 & 57.9 & 64.1 & 60.1 & 62.8 \\
NRC & 64.0 & 70.9 & 69.4 & 63.7 & 68.3 & 69.3 & 62.8 & 68.8 & 58.9 & 67.1 & 61.4 & 63.7 \\
3C-GAN & 64.8 & 68.5 & 69.1 & 61.4 & 68.3 & 67.9 & 59.9 & 62.5 & 57.0 & 61.0 & 58.3 & 60.7 \\
G-SFDA & 62.4 & 69.9 & 67.8 & 63.1 & 68.6 & 67.9 & 61.2 & 66.8 & 57.8 & 63.3 & 60.6 & 62.6 \\
\midrule
DANN & 62.4 & 67.9 & 68.0 & 60.9 & 65.2 & 67.3 & 59.3 & 60.9 & 57.0 & 60.4 & 57.3 & 59.5 \\
CORAL & 63.8 & 68.7 & 66.3 & 62.1 & 64.9 & 66.8 & 56.9 & 60.1 & 55.9 & 58.6 & 56.9 & 58.7 \\
\midrule
SPDNet & 56.3 & 62.0 & 62.5 & 51.5 & 60.9 & 59.4 & 48.2 & 44.8 & 52.8 & 47.3 & 52.1 & 49.9 \\
SPD-GNN & 57.0 & 64.9 & 66.0 & 55.4 & 61.7 & 63.0 & 52.0 & 48.0 & 52.7 & 52.0 & 53.0 & 50.6 \\
\midrule
UDA-GCN & 62.7 & 71.2 & 69.9 & 61.9 & 67.3 & 68.6 & 62.5 & 68.6 & 58.7 & 65.5 & 60.8 & 63.1 \\
StruRW & 64.2 & 70.6 & 70.2 & 64.5 & 69.8 & 70.4 & 62.8 & 69.4 & 58.8 & 66.7 & 61.5 & 64.0 \\
\midrule
BrainGNN & 63.8 & 69.2 & 68.9 & 63.2 & 68.4 & 68.5 & 52.0 & 51.2 & 54.6 & 54.2 & 54.4 & 54.6 \\
A-GCL & 63.5 & 70.3 & 69.9 & 62.5 & 68.8 & 69.3 & 55.8 & 56.4 & 55.0 & 53.2 & 54.7 & 53.7 \\
\midrule
\textbf{Ours} & \textbf{66.8} & \textbf{72.1} & \textbf{71.8} & \textbf{65.9} & \textbf{71.0} & \textbf{71.4} & \textbf{68.1} & \textbf{77.8} & \textbf{61.2} & \textbf{74.2} & \textbf{65.8} & \textbf{70.2} \\
\bottomrule
\end{tabular}
}
\end{table*}

\textbf{Baselines.} We compare with state-of-the-art methods across five categories: Source-Free DA (SHOT~\cite{liang2020we}, NRC~\cite{yang2021exploiting}, 3C-GAN~\cite{rahman20213c}, G-SFDA~\cite{yang2021generalized}), Traditional DA (DANN~\cite{ganin2016domain}, CORAL~\cite{sun2016deep}), Graph DA (UDA-GCN~\cite{wu2020unsupervised}, StruRW~\cite{liu2023structural}), and Source-Only baselines (SPDNet~\cite{huang2018building}, SPD-GNN~\cite{cheng2023spd}, BrainGNN~\cite{li2021braingnn}, A-GCL~\cite{zhang2023gcl}). See Supplementary Material B.2 for detailed method descriptions.

\textbf{Implementation Details.} Our framework was implemented using PyTorch on a server equipped with 8 NVIDIA RTX 4090 GPUs. We employed the Graph Isomorphism Network (GIN) as the backbone encoder for learning node representations. We utilized the Adam optimizer with a learning rate of $1 \times 10^{-3}$ and weight decay of $5 \times 10^{-4}$. The training process consisted of 100 epochs, with a batch size of 64 for both source and target domains. For the prototype learning, we set the number of prototypes $K=4$ per class based on validation performance. The trade-off hyperparameters were determined via grid search, with optimal values $\lambda_1=0.1$, $\lambda_2=0.01$, $\lambda_3=1.0$, and $\lambda_4=0.05$. To ensure fair comparison, all baselines were trained using their official implementations with optimized hyperparameters for our datasets.

\subsection{Cross-Site Adaptation Results}

\textbf{Single-Source Adaptation.} Tables~\ref{tab:abide_single_source} and~\ref{tab:mdd_single_source} present single-source adaptation results on ABIDE and REST-meta-MDD. Here, we select one large site as the labeled source, adapting to remaining sites individually. BrainRiem consistently outperforms all baselines across all source-target pairs. On ABIDE, it achieves 5.8\% average improvement over the best baseline StruRW, with notable gains on NYU and UM sources. On REST-meta-MDD, BrainRiem maintains superiority across all 20 transfers, excelling in cross-age adaptations (e.g., S25$\to$S14: 66.8\% vs. 62.4\%) and validating age-invariant biomarker capture. This demonstrates our manifold-aware prototypes effectively transfer knowledge to diverse targets despite scanner and demographic variations.

\textbf{Leave-One-Site-Out Adaptation.} Table~\ref{tab:loso} presents leave-one-site-out (LOSO) results on representative sites (full results in Supplementary Material B.5). On ABIDE, our method achieves 71.4\% average accuracy across all 15 sites, outperforming the best source-free baseline NRC (69.3\%) and graph adaptation method StruRW (70.4\%). On REST-meta-MDD, BrainRiem achieves 70.2\% average accuracy across 23 sites, significantly outperforming NRC (63.7\%) by 6.5\% and StruRW (64.0\%) by 6.2\%. The substantial gains are particularly evident on sites with extreme demographics (e.g., S9: 77.8\%, S14: 74.2\%), where traditional methods struggle due to severe age and gender shifts. Consistent superiority demonstrates our manifold-aware prototype learning's robustness in disentangling pathological patterns from site-specific confounds.

\subsection{Ablation Study}

To validate the contribution of each component in our BrainRiem framework, we conduct comprehensive ablation experiments on the REST-meta-MDD dataset under the leave-one-site-out setting.

\begin{figure}[t]
\centering
\includegraphics[width=\columnwidth]{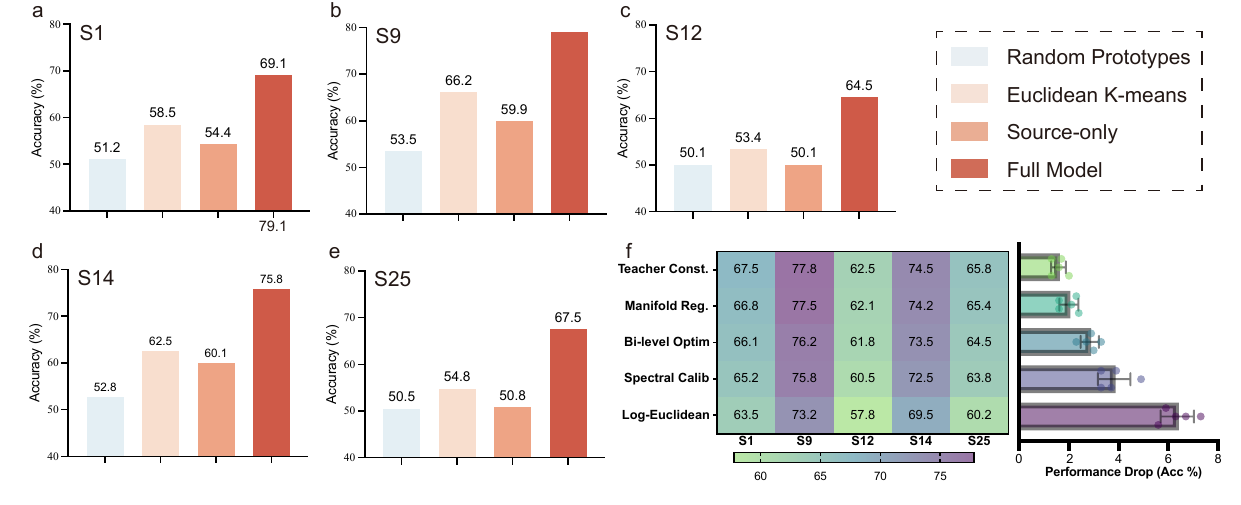}
\caption{Ablation analysis (REST-meta-MDD, LOSO). \textbf{(a-e)} Prototype initialization comparison across five sites; Riemannian-aware initialization outperforms baselines. \textbf{(f)} Component importance: Log-Euclidean mapping removal causes largest performance drop.}
\label{fig:ablation}
\end{figure}

\textbf{Impact of Prototype Initialization.} As shown in Figure~\ref{fig:ablation}(a-e), we compare different prototype initialization strategies on the REST-meta-MDD dataset. Random Prototypes yields near-chance performance ($\sim$50\%), indicating that uninformed initialization fails to capture meaningful class structure in depression diagnosis. Euclidean K-means improves upon this but still underperforms due to its neglect of the Riemannian geometry of brain connectivity matrices. Source-only (No Adapt.) transfers the source model directly without adaptation, showing moderate performance degradation caused by cross-site distribution shift. Our full BrainRiem framework with geometry-aware prototype initialization achieves the best performance across all MDD target sites.

\textbf{Contribution of Individual Components.} Figure~\ref{fig:ablation}(f) presents a detailed analysis of each component's contribution on the REST-meta-MDD dataset. The left heatmap visualizes the performance of each ablated variant across MDD sites, while the right chart quantifies the performance drop when each component is removed. Notably, removing the Log-Euclidean mapping (w/o Log-Euclidean) causes the most significant performance degradation ($\sim$5.7\% drop), confirming that proper Riemannian geometry handling is crucial for depression-related brain connectivity analysis. The spectral calibration loss ($\mathcal{L}_{spectral}$) and manifold regularization ($\mathcal{L}_{manifold}$) also contribute substantially, stabilizing cross-site adaptation and preserving intrinsic data structure. The teacher-guided prototype supervision term and bi-level optimization provide complementary benefits, ensuring consistent knowledge transfer and balanced prototype-classifier learning.

\subsection{Hyperparameter Analysis}
We conducted a sensitivity analysis to investigate the impact of key hyperparameters on the adaptation performance, specifically the number of prototypes $K$ and the regularization weights.

\begin{figure*}[t]
\centering
\includegraphics[width=\textwidth]{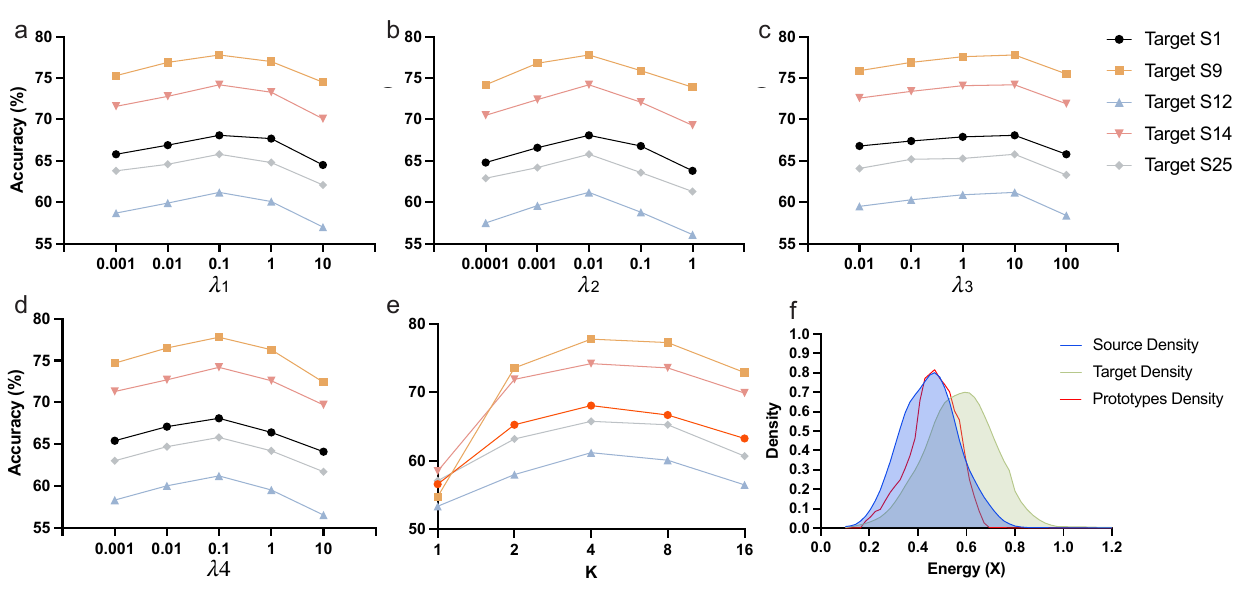}
\caption{Hyperparameter and spectral analysis (REST-meta-MDD, LOSO). \textbf{(a-d)} Sensitivity of $\lambda_1$-$\lambda_4$ across five sites; framework shows robustness. \textbf{(e)} Optimal $K=4$ prototypes. \textbf{(f)} Dirichlet Energy: learned prototypes align with source/target distributions, preventing high-frequency artifacts.}
\label{fig:hyper_combined}
\end{figure*}

\textbf{Number of Prototypes ($K$).} The number of prototypes per class determines the granularity of the knowledge transfer. We varied $K$ within $\{1, 2, 4, 8, 16\}$. As shown in Figure~\ref{fig:hyper_combined}(e), performance initially improves as $K$ increases, as more prototypes can better capture the intra-class diversity of brain functional patterns. However, setting $K$ too large (e.g., $>8$) led to a slight performance drop, likely due to the prototypes capturing site-specific noise rather than shared diagnostic features. We observed that $K=4$ yields the most robust performance across both datasets.

\textbf{Regularization Weights.} We analyzed the sensitivity of $\lambda_1$ (manifold), $\lambda_2$ (spectral), $\lambda_3$ (teacher-guided supervision), and $\lambda_4$ (target entropy). As shown in Figure~\ref{fig:hyper_combined}(a-d), the framework demonstrated relative stability within an order of magnitude (see Supplementary Material B.3 for extended analysis).

\subsection{Privacy Analysis}
\label{sec:privacy}
To evaluate leakage under the evaluated attacks, we conducted a quantitative privacy audit. We implemented two standard privacy attacks: Re-identification Attack and Membership Inference Attack (MIA), using the NYU site data ($N=184$).

\textbf{Setup.} We partitioned NYU data (50/50 split) for re-identification and membership inference attacks (see Supplementary Material B.4 for protocols).

\begin{figure}[t]
\centering
\includegraphics[width=0.5\columnwidth]{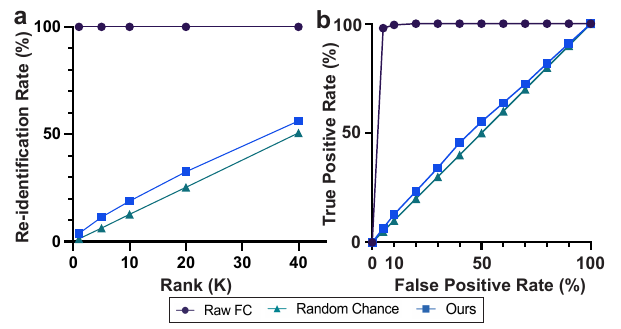}
\caption{Privacy auditing (NYU). \textbf{(a)} Re-identification: raw FC achieves 100\% Rank-1, BrainRiem reduces to 3.8\% (random-level). \textbf{(b)} Membership inference: raw AUC=0.99, BrainRiem AUC=0.53 (near-random).}
\label{fig:privacy}
\end{figure}

\textbf{Results.} Figure~\ref{fig:privacy} presents the privacy auditing results. For re-identification defense (Fig.~\ref{fig:privacy}a), transmitting raw FC matrices leads to 100\% Rank-1 identification, confirming that raw connectivity patterns act as identifiable fingerprints. In contrast, BrainRiem prototypes suppress the Rank-1 identification rate to 3.8\%, statistically comparable to random guessing ($1/N \approx 1.3\%$). For MIA defense (Fig.~\ref{fig:privacy}b), the baseline (raw transmission) exhibits AUC of 0.99, indicating severe privacy leakage. Conversely, the attack on BrainRiem prototypes yields AUC of 0.53, near the theoretical random baseline of 0.50. These results indicate reduced leakage under the evaluated attacks while preserving class-discriminative information.

\section{Discussion}

\subsection{Biological Interpretability of Learned Prototypes}

\begin{figure}[t]
    \centering
    \includegraphics[width=0.9\textwidth]{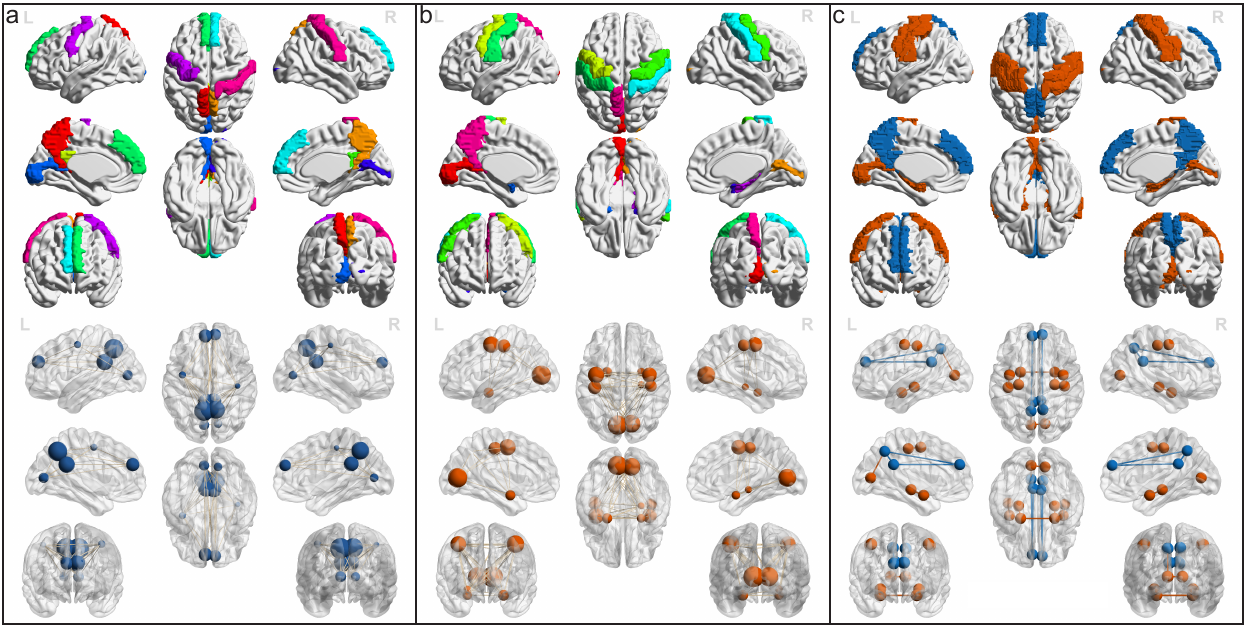}
    \caption{Learned brain prototypes (ABIDE). \textbf{(a)} TD: DMN-dominated ``Rich-Club'' organization. \textbf{(b)} ASD: sensory-motor and limbic dominance. \textbf{(c)} Differential (ASD$-$TD): hyper-connectivity in sensory-motor, hypo-connectivity in DMN.}
    \label{fig:proto_abide}
\end{figure}

To verify the biological validity of learned prototypes, we visualized their topological structures in Figure~\ref{fig:proto_abide}. We compute the Log-Euclidean Mean $\bar{\mathbf{A}}_{geo} = \text{Exp}(\frac{1}{K}\sum_{k=1}^K \text{Log}(\mathbf{A}_k))$ to avoid the ``swelling effect'' from naive averaging. The TD prototype exhibits a characteristic ``Rich-Club'' organization dominated by the Default Mode Network (DMN), while the ASD prototype shows a distinct shift towards primary sensory and limbic systems, aligning with the ``sensory dominance'' hypothesis in autism research. The differential network (Figure~\ref{fig:proto_abide}c) reveals \textit{hyper-connectivity} in sensory-motor loops and \textit{hypo-connectivity} in DMN core regions, which provides evidence for the ``DMN-underconnectivity'' hypothesis (see Supplementary Material C.1--C.2 for complete analysis).

\subsection{Spectral Calibration and Privacy Analysis}
As shown in Figure~\ref{fig:hyper_combined}(f), the target domain exhibits a distributional shift towards higher Dirichlet energy, indicating site-specific scanner noise. Our prototypes maintain a spectral profile aligned with the source domain, confirming that $\mathcal{L}_{spectral}$ enforces biological plausibility (see Supplementary Material C.3). Additionally, our source-free paradigm transmits only compact prototypes ($K \times 2 = 8$ matrices per task), which reduces leakage under the evaluated attacks. Privacy auditing (Section~\ref{sec:privacy}) indicates reduced functional connectivity ``fingerprinting'' by acting as a geometric low-pass filter, retaining diagnostic patterns while smoothing individual-specific deviations.

\subsection{Theoretical Validation}
The ablation study (Figure~\ref{fig:ablation}(f)) confirms that removing the Log-Euclidean mapping substantially degraded performance, validating that SPD manifold structure is essential for knowledge transfer. Euclidean K-means prototypes exhibited dense, noisy connectivity manifesting the ``swelling effect,'' while our manifold-aware prototypes preserved sparse, biologically interpretable structures matching real fMRI's hierarchical modular architecture.

\section{Conclusion}
We presented BrainRiem, a source-free domain adaptation framework for cross-site brain network diagnosis. By learning compact brain prototypes on the Riemannian manifold via bi-level optimization, our method enables robust knowledge transfer without accessing source data during adaptation and reduces leakage under the evaluated attacks. Comprehensive experiments on ABIDE and REST-meta-MDD demonstrated state-of-the-art performance across diverse scanner configurations, while the learned prototypes exhibited clear biological interpretability and spectral consistency with real fMRI data. Beyond the current setting, BrainRiem provides a flexible foundation for future extensions, including the use of multi-scale or data-driven parcellations to reduce atlas dependency, manifold-aware modeling of dynamic functional connectivity to capture time-varying brain dynamics, and open-set domain adaptation on the SPD manifold to better reflect realistic clinical scenarios involving novel disease subtypes or comorbidities. Overall, BrainRiem offers a practical paradigm for leakage-aware multi-center neuroimaging collaboration, with potential applicability to other brain disorders and imaging modalities.

%
%
\bibliographystyle{splncs04}
\bibliography{ref}


\end{document}